\documentclass[journal,twoside,web]{ieeecolor}
\usepackage{etoolbox}
\makeatletter
\@ifundefined{color@begingroup}%
{\let\color@begingroup\relax
\let\color@endgroup\relax}{}%
\def\fix@ieeecolor@hbox#1{%
\hbox{\color@begingroup#1\color@endgroup}}
\patchcmd\@makecaption{\hbox}{\fix@ieeecolor@hbox}{}{\FAILED}
\patchcmd\@makecaption{\hbox}{\fix@ieeecolor@hbox}{}{\FAILED}
\usepackage{tmi}
\usepackage{cite}
\usepackage{amsmath,amssymb,amsfonts}
\usepackage{algorithmic}
\usepackage{graphicx}
\usepackage{textcomp}
\usepackage{hyperref}
\usepackage{multirow}
\usepackage{booktabs} 
\hypersetup{
hypertex=true,
colorlinks=true,
linkcolor=red,
anchorcolor=red,
citecolor=blue}
\def\BibTeX{{\rm B\kern-.05em{\sc i\kern-.025em b}\kern-.08em
    T\kern-.1667em\lower.7ex\hbox{E}\kern-.125emX}}
\begin{document}
\title{Medical Referring Image Segmentation via Next-Token Mask Prediction}
\author{Xinyu Chen, Yiran Wang, Gaoyang Pang, Jiafu Hao, Chentao Yue,\\ Luping Zhou, \IEEEmembership{Senior Member, IEEE}, and Yonghui Li, \IEEEmembership{Fellow, IEEE}
\thanks{Xinyu Chen, Yiran Wang, Gaoyang Pang, Jiafu Hao, Chentao Yue, Luping Zhou, and Yonghui Li are with the School of Electrical and Computer Engineering,
University of Sydney, Sydney, NSW 2006, Australia. (e-mail: \href{mailto:xinyu.chen@sydney.edu.au}{\textcolor{black}{xinyu.chen@sydney.edu.au}};
\href{mailto:ywan0327@uni.sydney.edu.au}{\textcolor{black}{ywan0327@uni.sydney.edu.au}};
\href{mailto:gaoyang.pang@sydney.edu.au}{\textcolor{black}{gaoyang.pang@sydney.edu.au}}; 
\href{mailto:jiafu.hao@sydney.edu.au}{\textcolor{black}{jiafu.hao@sydney.edu.au}}; 
\href{mailto:chentao.yue@sydney.edu.au}{\textcolor{black}{chentao.yue@sydney.edu.au}}; 
\href{mailto:luping.zhou@sydney.edu.au}{\textcolor{black}{luping.zhou@s\allowbreak ydney.edu.au}}; 
\href{mailto:yonghui.li@sydney.edu.au}{\textcolor{black}{yonghui.li@sydney.edu.au}}).\\
This work has been submitted to the IEEE Transactions on Medical Imaging for possible publication. Copyright may be transferred without notice, after which this version may no longer be accessible.
}}

\maketitle

\begin{abstract}
Medical Referring Image Segmentation (MRIS) involves segmenting target regions in medical images based on natural language descriptions. While achieving promising results, recent approaches usually involve complex design of multimodal fusion or multi-stage decoders. In this work, we propose NTP-MRISeg, a novel framework that reformulates MRIS as an autoregressive next-token prediction task over a unified multimodal sequence of tokenized image, text, and mask representations. This formulation streamlines model design by eliminating the need for modality-specific fusion and external segmentation models, supports a unified architecture for end-to-end training. It also enables the use of pretrained tokenizers from emerging large-scale multimodal models, enhancing generalization and adaptability. More importantly, to address challenges under this formulation—such as exposure bias, long-tail token distributions, and fine-grained lesion edges—we propose three novel strategies: (1) a Next-k Token Prediction (NkTP) scheme to reduce cumulative prediction errors, (2) Token-level Contrastive Learning (TCL) to enhance boundary sensitivity and mitigate long-tail distribution effects, and (3) a memory-based Hard Error Token (HET) optimization strategy that emphasizes difficult tokens during training. Extensive experiments on the QaTa-COV19 and MosMedData+ datasets demonstrate that NTP-MRISeg achieves new state-of-the-art performance, offering a streamlined and effective alternative to traditional MRIS pipelines.
\end{abstract}


\section{Introduction}
\label{sec:introduction}
\IEEEPARstart{M}{edical} Referring Image Segmentation (MRIS) involves segmenting the specific lesions described in a natural language. Compared with conventional medical image segmentation tasks \cite{pereira2016brain, gu2019net, fan2020inf, gu2020net} that handle only a fixed set of categories, MRIS offers greater flexibility by allowing the segmentation of arbitrary anatomical structures, lesions, or abnormalities described in free-text form \cite{li2023lvit}. This capability requires Artificial Intelligence (AI) to have a comprehensive understanding and alignment between diverse medical terminology and radiological images, which can be leveraged in clinical scenarios, such as AI-assisted diagnosis.

Some approaches use traditional single-modal pre-trained image or text backbones to extract features \cite{li2023lvit, zhong2023ariadne, ye2024enabling, guo2024common} such as incorporate textual prompts during the encoder stage to guide the segmentation network \cite{li2023lvit}. Others apply language guidance in the decoder stage \cite{zhong2023ariadne} or develop self-guided segmentation frameworks that iterate between vision and language processing \cite{ye2024enabling}. Benefiting from advances in cross-attention mechanisms \cite{lee2023text}, UNet-based architectures have also been extended for MRIS, achieving strong performance in recent studies \cite{guo2024common}. The emergence of large-scale vision-language models like Contrastive Language-Image Pretraining (CLIP)\cite{radford2021learning} has further spurred interest due to their impressive generalization capability. CLIP’s text encoder has been used to learn robust feature representations for medical images \cite{chen2024pcnet, kunhimon2024language}, and custom decoders have been designed to exploit CLIP’s rich semantic space in the medical domain \cite{chen2024causalclipseg}. However, current MRIS models often require specially designed fusion modules or rely on dedicated decoders or external segmentation components (e.g., SAM \cite{koleilat2024medclip}), leading to overly complex systems.

Recently, the Visual Autoregressive (VAR) modeling paradigm \cite{tian2024visual} has provided a conceptually simple and powerful alternative for vision tasks by unifying tasks as sequence predictions. Nevertheless, achieving effective multimodal fusion within a VAR framework remains a significant challenge. Next-Token Prediction (NTP) offers a unified approach to multimodal tasks by tokenizing images and text in a discrete space, and then predicting subsequent tokens in an autoregressive manner \cite{wang2024emu3}. Training on diverse multimodal token sequences can achieve effective vision-language understanding \cite{yue2024object}. This emerging NTP paradigm presents a promising opportunity to simplify MRIS models and eliminate the need for complex task-specific architecture.

\begin{figure}[t]
\centerline{\includegraphics[width=0.9\columnwidth]{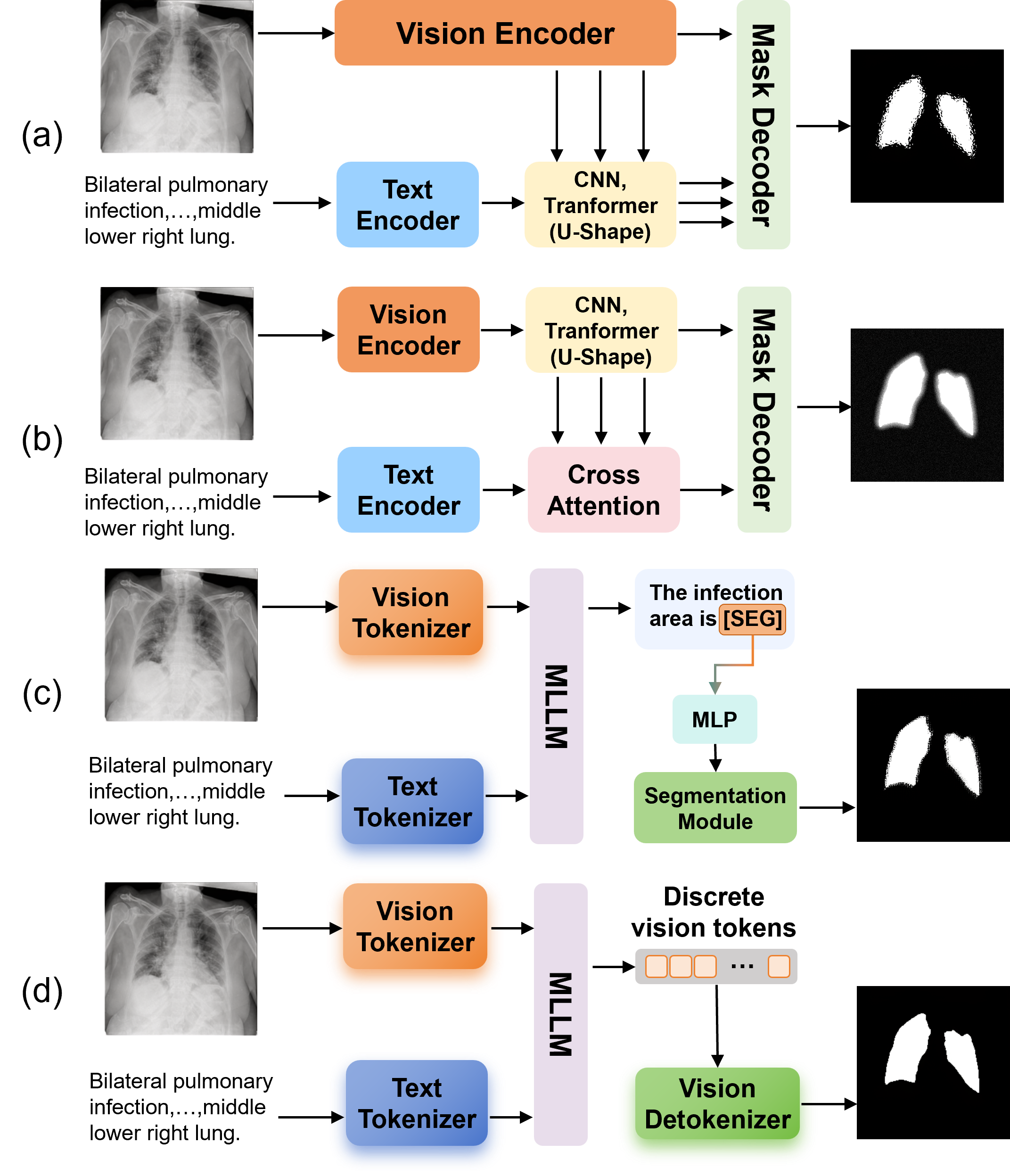}}
\caption{Comparison of different models for MRIS. (a) Models that integrate additional parallel U-shape architecture to aligns and fuse text features and vision features\cite{li2023lvit, huang2024cross}. (b) Dual-branch fusion architectures that apply cross attention to align and fuse text features and vision features \cite{hu2023beyond, ouyang2024lsms}. (c) MLLM-based models that align multimodal features and use embedded representations as masks for decoding\cite{hu2024lga, koleilat2024medclip}. \textbf{(d) Ours:} a unified MLLM-based framework that aligns features and directly uses visual tokens as mask inputs to a detokenizer.}
\label{Fig1_model_comparison}
\end{figure}

However, applying NTP to MRIS introduces its own difficulties. In an autoregressive model trained with teacher forcing, there is a mismatch between training and inference known as \emph{exposure bias} \cite{bachmann2024pitfalls}. During training the model sees ground-truth context tokens, but at inference it must rely on its own predicted tokens. As a result, early prediction errors can compound and propagate, leading to significant error accumulation. Moreover, representing a medical image segmentation mask as a sequence of tokens can exacerbate long-tail distribution problems \cite{zhang2023deep, jiang2022simple}: common tokens (representing large regions) dominate the training data while rare tokens (e.g., lesion edges or small abnormalities) are underrepresented. Lesion segmentation is a fine-grained task requiring exceptional precision at region edges, and the imbalanced distribution of lesion vs. background pixels further complicates learning. The comparison of existing methods for MRIS is summarized in Fig.~\ref{Fig1_model_comparison}.

To address the above challenges, we propose NTP-MRISeg, a novel framework that reformulates MRIS as an autoregressive next-token mask prediction task. Our method uses a pure Transformer architecture that predicts segmentation masks token-by-token, eliminating the need for diffusion processes or composite pipelines (see Fig.~\ref{Fig2_model_overall} for an overview). Our contributions are summarized as follows:

\textbf{1) Unified NTP-based Framework:} First, we propose a unified autoregressive formulation for MRIS that tokenizes medical images, referring expressions, and segmentation masks into a single multimodal sequence, enabling segmentation through next-token prediction. This architecture removes the need for handcrafted modality-specific fusion or separate decoding modules, offering a streamlined and extensible framework that naturally supports end-to-end training and integration with large-scale pretrained tokenizers. Furthermore, to mitigate exposure bias, we introduce a Next-k Token Prediction (NkTP) strategy that improves sequence consistency by predicting future $k$ tokens during training.

\textbf{2) Robust Token-level Contrastive Learning:} Second, we propose a contrastive learning scheme at the token level (TCL), which explicitly pushes the model to separate rare tokens (like lesion edges) from nearby repeated or background tokens. This encourages the model to make fine-grained distinctions between similar tokens, enhancing its sensitivity to lesion edges and addressing the long-tail distribution of mask tokens.

\textbf{3) Hard Error Token Optimization:}  Third, we introduce a memory-driven mechanism (HET) that tracks historically mispredicted tokens across training epochs, ranks their difficulty, and uses them as hard negatives in contrastive learning. This targeted emphasis on challenging tokens improves the model’s ability to recover from persistent prediction errors and enhances segmentation precision in difficult lesion regions.

\textbf{4) State-of-the-Art Performance:} Fourth, our approach achieves new state-of-the-art results on both QaTa-COV19 and MosMedData$+$ datasets, demonstrating superior accuracy and robustness across modalities.

\section{Related Work}
\subsection{Referring Segmentation of Medical Images}
Referring Image Segmentation (RIS) is a task of segmenting the target region in images based on the given natural language description. Early works on RIS (in general computer vision) \cite{hu2016segmentation, li2018referring, liu2017recurrent} explored concatenating visual features from Convolutional Neural Networks (CNNs) and language features from Recurrent Neural Networks (RNNs), followed by convolutional fusion, to generate the segmentation mask. In the medical domain, RIS techniques can facilitate AI-assisted diagnosis by enabling interactive segmentation based on radiologists’ descriptions. With the success of attention mechanisms in multimodal learning, researchers began incorporating cross-attention into medical segmentation networks. For instance, some methods \cite{ye2024enabling} integrate textual context into a UNet-based architecture \cite{ronneberger2015u} via cross-attention to perform MRIS.

The breakthrough of Transformers \cite{vaswani2017attention} in computer vision has made them increasingly dominant in MRIS. Hybrid CNN–Transformer frameworks were introduced to merge medical image and text features more effectively \cite{li2023lvit, huang2024cross, guo2024common}. LViT \cite{li2023lvit} employed a pixel-level attention mechanism to enhance local feature details and align multimodal representations. TGCAM \cite{guo2024common} combined standard cross-attention with iterative text feature enhancement to improve interaction between modalities. TPP \cite{yuan2025text} extracted sequential dependencies from time-series medical scans and their reports to achieve sequence-level referring segmentation. Unlike DMMI \cite{hu2023beyond}, which only reconstructed randomly erased phrases to enforce cross-modal consistency, RecLMIS \cite{huang2024cross} performed a bidirectional visual-text conditioned reconstruction to explicitly capture fine-grained interactions. Conventional multimodal segmentation approaches based on CNN encoders, such as ConViRT\cite{zhang2022contrastive} and TGANet\cite{tomar2022tganet}, struggled to fully leverage textual information due to limited cross-modal fusion. In summary, these methods have effectively bridged modality gaps and improved MRIS performance. However, achieving MRIS with the above model architectures often requires complex combinations of modules, motivating the search for a more streamlined approach.

\begin{figure*}[t]
\centerline{\includegraphics[width=2\columnwidth]{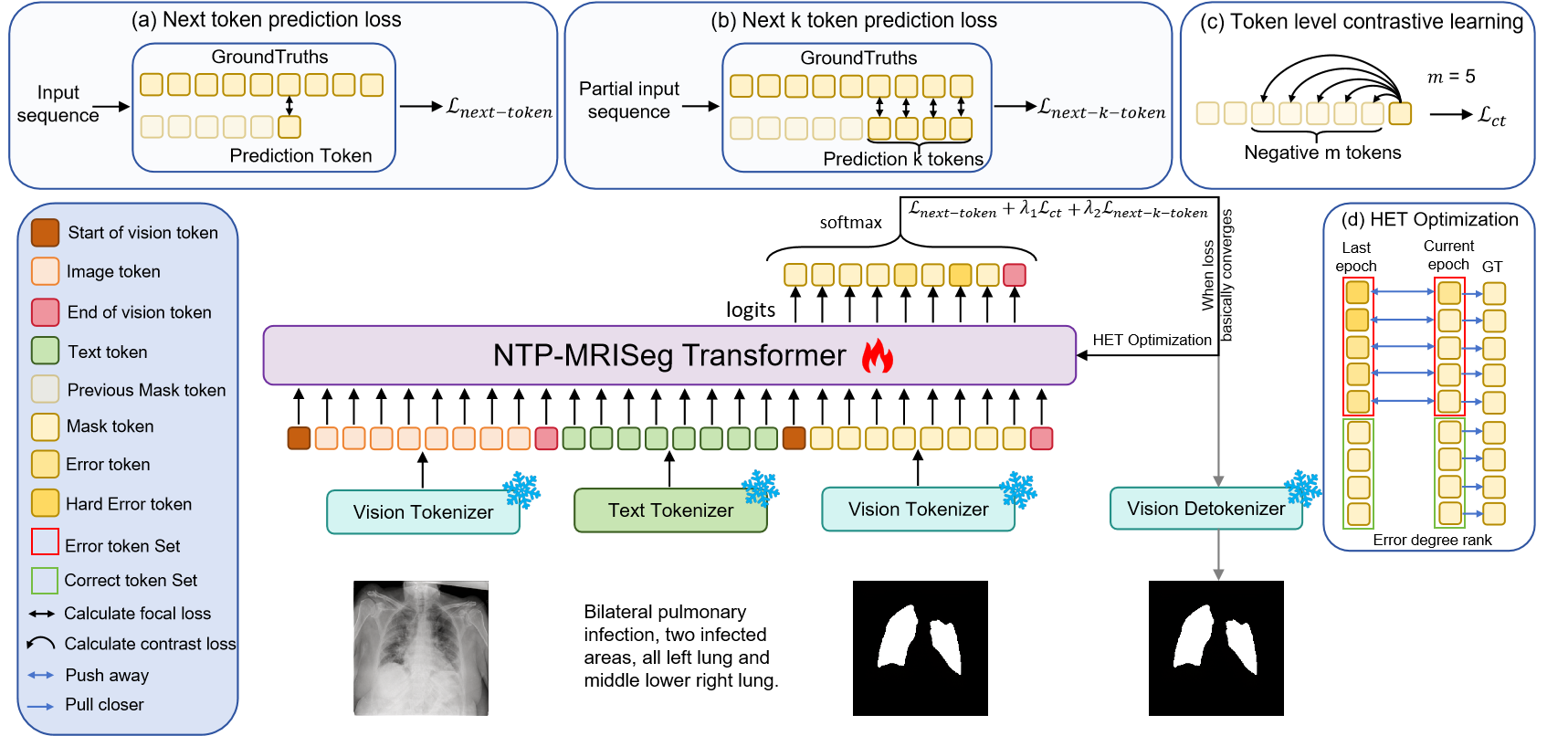}}
\caption{Overall framework of NTP-MRISeg. (a) Mechanism of NTP: the model predicts each token in the sequence based on preceding tokens, with loss calculated by comparing predicted tokens against Ground Truth (GT) labels. (b) Mechanism of NkTP: the model simultaneously predicts $k$ consecutive tokens based on preceding tokens, with loss calculated across all $k$ predicted tokens against their corresponding GT. (c) Mechanism of TCL: each token uses its corresponding GT as the positive sample and the preceding m predicted tokens ($m=5$ in this example) as negative samples for contrastive learning. (d) Mechanism of HET optimization: error tokens from the previous epoch are ranked by deviation from ground truth, with the most challenging errors selected to push predictions away from historical error tokens while pulling them closer to GT.}
\label{Fig2_model_overall}
\end{figure*}

\subsection{Multimodel Large Language Model}
Large Language Models (LLMs) have demonstrated exceptional reasoning capabilities, and recent research extends these abilities to vision tasks via multimodal LLMs (MLLMs). For example, BLIP-2 \cite{li2023blip}, mPLUG-OWL \cite{ye2023mplug}, LLaVA \cite{liu2023visual}, and related frameworks \cite{zhu2023minigpt,wang2023visionllm,lai2024lisa,rasheed2024glamm} integrate visual encoders with LLMs to enable tasks like visual question answering and referring image understanding. These MLLMs typically use a pre-trained LLM to process textual inputs and a vision backbone (CNN or ViT) to encode images, bridging the two modalities through learned projection layers or attention. They have achieved impressive results on general multimodal benchmarks, demonstrating the potential of unified vision-language reasoning.

In the medical imaging domain, there are emerging efforts to adapt MLLMs for tasks such as clinical image interpretation and report generation. For instance, CLIP \cite{radford2021learning} is a pioneering vision-language model that has been applied to medical images to bridge modality gaps in segmentation. Causal-CLIPSeg \cite{chen2024causalclipseg} builds on CLIP by adding a tailored cross-modal decoding component to better utilize CLIP’s semantic space for medical segmentation. PCNet \cite{chen2024pcnet} leverages CLIP features with attention mechanisms to establish relationships between anatomical categories defined by clinicians, improving segmentation performance. While these large pre-trained models provide powerful semantic representations, directly applying general-purpose MLLMs to MRIS is non-trivial. The medical domain has specialized terminology and fine-grained diagnostic details that generic models may not capture, and segmentation requires precise localization beyond the typical output of an LLM. In summary, MLLM-based approaches show promise in combining visual and textual understanding, but they have yet to fully meet the fine-grained, high-precision requirements of MRIS. This gap motivates our task-specific approach, which uses an autoregressive segmentation model with optimizations tailored for medical images and descriptions.

\section{Method}
Recently, VAR\cite{tian2024visual}, as a new paradigm, has demonstrated its powerful performance in visual generation. In this context, Emu3\cite{wang2024emu3} tokenizes images and text in a discrete space as tokens and employs a pure transformer-base model using only NTP on diverse multimodal sequences, simplifies multimodal designs. These methods showcase NTP's promising potential in multimodal and generation tasks and motivate the development of our NTP-MRISeg detailed in the following. 

\subsection{NTP-MRISeg Framework}
Our proposed NTP-MRISeg provides a unified framework for MRIS tasks based solely on next-token prediction, completely eliminating the need for compositional methods, as shown in Fig. \ref{Fig2_model_overall}. We tokenize medical images and pathology descriptions into a discrete space and jointly train a single transformer from scratch on a mixture of multimodal sequences. To ensure optimal model adaptation to MRIS tasks, we carefully design NkTP to compensate for exposure bias between training and inference, TCL to address the long-tail distribution problem through contrastive learning against preceding m tokens, and HET to specifically optimize challenging difficult tokens. Next, we will elaborate the structure details of our proposed framework.
\begin{figure}[t]
\centerline{\includegraphics[width=\columnwidth]{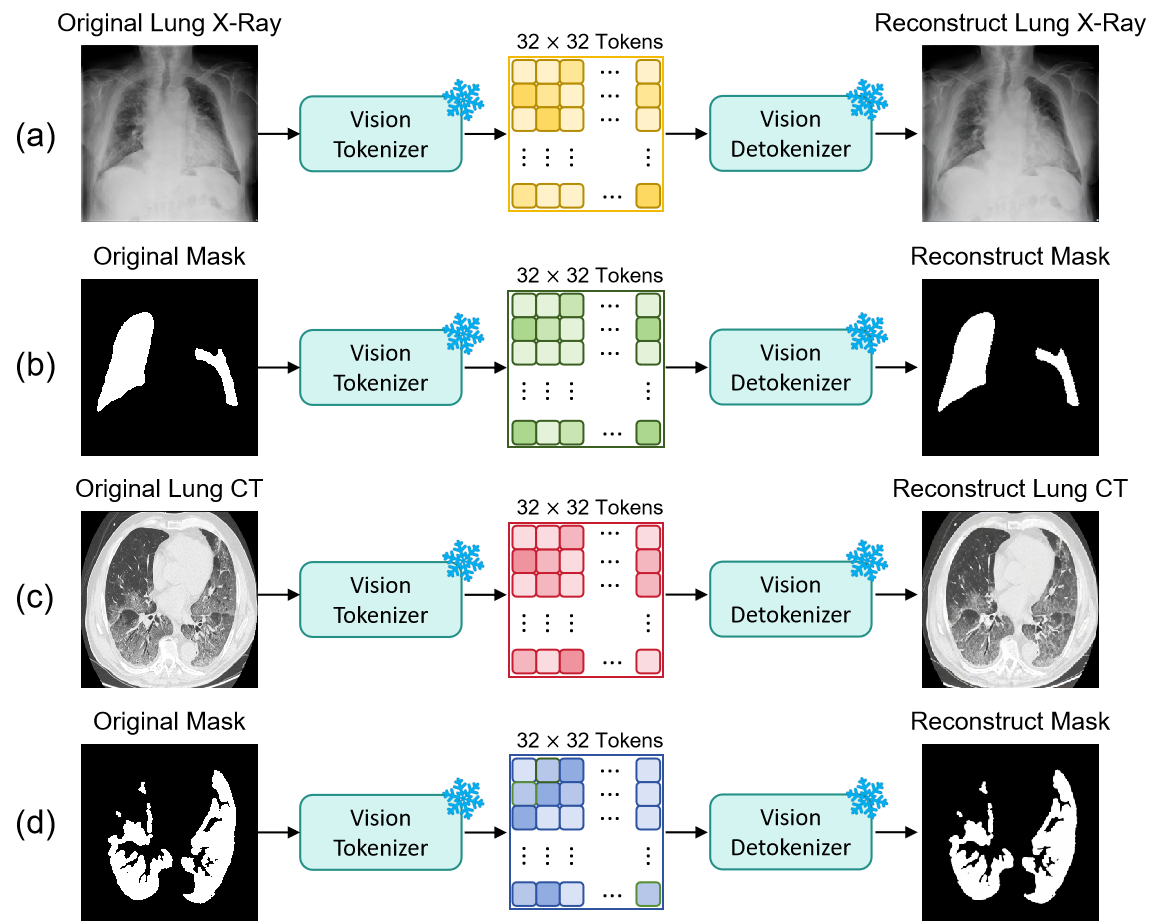}}
\caption{Visualization of original and reconstructed medical images and masks using the Emu3 SBER-MoVQGAN tokenizer. (a) Original lung X-ray image, (b) Corresponding segmentation mask, (c) Original lung CT image, (d) Corresponding segmentation mask. Each image and mask is tokenized and then reconstructed from discrete tokens. The preservation of structural and boundary details demonstrates the tokenizer’s suitability for MRIS.}
\label{Fig_recon}
\end{figure}

\subsubsection{Vision and Text Tokenizer}
We employ the vision tokenizer based on Emu3 SBER-MoVQGAN\cite{razzhigaev2023kandinsky}, which achieves 8$\times$8 spatial compression and supports arbitrary spatial resolutions. Specifically, a 256$\times$256 medical image is encoded into a 32$\times$32 grid of discrete tokens, each selected from a codebook of size 32,768. To demonstrate the suitability of this general-purpose tokenizer for medical imaging tasks, Fig. \ref{Fig_recon} shows examples of medical images and corresponding segmentation masks that are first tokenized and then reconstructed from tokens. The reconstructed results confirm that critical structural details and edge textures are well preserved, validating the effectiveness of the Emu3 tokenizer for MRIS. And we use Qwentokenizer\cite{bai2023qwen} for medical descriptions.

\subsubsection{Multimodel Data Preparation}
To implement the MRIS task, we define a unified multimodal data format. Unlike diffusion models that depend on external text encoders, NTP-MRISeg integrates text-conditioned information with medical images. Following image resizing to a fixed dimension, we employ visual and text tokenizers to generate corresponding visual and text tokens. Subsequently, we incorporate four special tokens to seamlessly combine textual and visual data, creating a document-like input structure for the training process. The resulting training data follows this structure:
\begin{center}
    \text{[BOS] [medical images] \{descriptions\} [seg masks] [EOS],}
\end{center}
where [BOS] and [EOS] are the original special tokens in the text tokenizer. [medical images] and [seg masks] follow the following format:
\begin{center}
    \text{[SOV] \{meta text\} [SOT] \{vision tokens\} [EOV],}
\end{center}
where [SOV] and [EOV] indicate the start and end of vision input, [SOT] mark the start of vision tokens. The \{meta text\} contains the resolution information for images. Through the token sequence construction incorporating medical images, medical descriptions, and segmentation masks, the model naturally adapts to the MRIS task.

\subsubsection{Model Architecture}
\label{sec:model}
The NTP-MRISeg model employs a transformer-based architecture fundamentally rooted in established LLMs, specifically following the architectural principles of Llama-2 while incorporating the multimodal design paradigm from Emu3. The key innovation involves extending the traditional text embedding layer to seamlessly integrate discrete visual tokens, enabling unified processing of both textual and visual information within a unified framework. 

The model incorporates three key architectural optimizations. RMSNorm is employed for computational efficiency and training stability, eliminating mean-centering operations to reduce overhead during large-scale multimodal training. Grouped Query Attention (GQA) balances efficiency and expressiveness by enabling query heads to share key-value pairs, reducing memory consumption while preserving cross-modal modeling capabilities. The SwiGLU activation function provides smoother gradients and enhanced information flow for diverse multimodal feature representations.

Since both visual and textual signals in NTP-MRISeg are fully converted into discrete tokens, we employ focal loss based on standard cross-entropy loss to train the next token prediction task, which naturally addresses data imbalance issues as Fig. \ref{Fig2_model_overall}\textcolor{red}{a}.
Given an image $I$, we first tokenize it into a sequence of $N$ discrete tokens sequence $\mathbf{i} \triangleq (i_{1}, . . . , i_{N})$. Standard autoregressive modeling typically adopts a fixed left-to-right factorization:
\begin{equation}
    p(\mathbf{i}) = \prod_{n=1}^{N} p(i_n | i_{<n}),
\end{equation}
where $i_{<n}$ denotes all tokens preceding $i_{n}$ and the conditional probability $p(i_n|i_{<n})$ can be described as:
\begin{align}
        p(i_n|i_{<n}) &= \frac{\exp(h_n^\top E_{i_n})}{\sum_{\hat{i}_n \in S} \exp(h_n^\top E_{\hat{i}_n})} \nonumber \\
        &=\frac{1}{1\!\!+\!\!\sum_{\hat{i}_n \in S, \hat{i}_n \neq i_{n}} \exp(h_n^\top E_{\hat{i}_n}\!\!-\!h_n^\top E_{i_n})},
        \label{focal_hidden}
\end{align}
where $h_{n}$ denotes the model hidden state at position $n$, $i_{n}$ denotes the $n$-th token in the sequence, $\hat{i}_n$ denotes each candidate token in the vocabulary $S$. $E_{\hat{i}_n}$ is the embedding of each candidate token $\hat{i}_{n}$, $E_{i_n}$ is the embedding of Ground Truth (GT) token $i_{n}$, and $S$ represents the vocabulary of all tokens. The loss for training the model to predict the $n$-th token $i_{n}$ given the preceding context $i_{<n}$ can be described as:
\begin{equation}
    \mathcal{L}_{\text{next-token}} = -\sum_{n=1}^{N} \alpha (1-p(i_n|i_{<n}))^{\gamma} \log p(i_n|i_{<n}),
    \label{focal_loss}
\end{equation}
where $\alpha \in (0,1]$ is the balancing factor and $\gamma \in [0, + \infty)$ is the focusing parameter. NTP-MRISeg inherits the robust vision-language understanding capabilities of autoregressive models. However, the deterministic generation requirements of MRIS are more sensitive to cumulative errors caused by exposure bias. To address this challenge, we propose a novel auxiliary training strategy, i.e., the NkTP strategy, in the next subsection.

\subsection{Next-k Token Prediction Strategy}
\label{sec:NkTP}
Autoregressive inference represents a form of inference-time next-token prediction where, to generate a response, we iteratively sample the next token. Most autoregressive models employ teacher-forced training, which constitutes a form of training-time next-token prediction. In this approach, instead of feeding the model its own output as input, the model receives prefixes of the GT response. This discrepancy between the predicted responses used during inference and the GT prefixes used during training prevents the model from learning to recover from its own errors during inference.

To mitigate the ``snowball'' in effect of the training-inference discrepancy on fine-grained MRIS, we intuitively extend the training strategy by incorporating NkTP as Fig. \ref{Fig2_model_overall}\textcolor{red}{b} alongside traditional next-token prediction:
\begin{equation}
    \mathcal{L}_{\text{next-k-token}} \!=\! -\!\sum_{n=1}^{N} \sum_{k=n}^{K} \alpha (1\!-\!p(i_k|i_{<n}))^{\gamma} \log p(i_k|i_{<n}),
    \label{next-k-loss}
\end{equation}
where $k$ represents the number of additional tokens predicted more than only the next one. This demonstrates that NkTP optimizes the sum of log probabilities for each token $i_{k+n}$ over all preceding contexts $i<n$ for $n<i\leq k$, unlike the standard autoregressive objective that only considers the immediately preceding context.

By incorporating the NkTP auxiliary prediction task alongside NTP during training, we provide the model with opportunities to learn accurate and consistent long-term predictions, thereby reducing cumulative error. This auxiliary training strategy enables the model to generate sequences that are consistent with both the immediate context and $k$ potential future contexts, capturing more complex dependencies and interactions between distant tokens, which results in richer and more expressive representations.

Although introducing NkTP substantially alleviates the cumulative error problem inherent in the training mechanism, challenges remain due to the binary characteristics of segmentation masks. The model frequently generates long sequences of tokens with minimal distinguishing features, leading to reduced sensitivity to token position changes and making the model prone to lazy predictions. Furthermore, this exacerbates the long-tail distribution problem, making it particularly challenging to address.

\subsection{Token-level Contrastive Learning}
\label{sec:TCL}
Sequences of similar tokens frequently appear in the token distribution of segmentation masks. The current loss function does not impose additional penalties on the previously abundant negative and insignificant tokens, which causes the model to develop ``inertia" for subsequent inference. This makes the model insensitive to abrupt changes in foreground and background edge tokens, resulting in repetition problems. TCL provides an effective approach to enhance the model's sensitivity to token variations and further mitigate the long-tail distribution problem. 

According to \eqref{focal_hidden},  focal loss is applied to train the model by contrasting label tokens $i_{n}$ (positive samples) against the non-label tokens $\hat{i}_n \in S, \hat{i}_n \neq i_{n}$ (negative and irrelevant samples).
To further encourage the model to focus on negative samples in more contextually relevant areas, the core principle of contrastive training is to promote positive (GT) tokens at each position while penalizing negative (incorrectly repeated) tokens and leaving other irrelevant tokens as Fig. \ref{Fig2_model_overall}\textcolor{red}{c}. In this case, we can design the conditional probability of TCL based on \eqref{focal_hidden} as follows:
\begin{align}\label{ct_condition}
        p_{ct}(i_n|i_{<n}) =\frac{1}{1 + \sum_{i^{-}_n \in S^{-}_{m}} \exp(h_n^\top E_{i^{-}_n}-h_n^\top E_{i_n})},
\end{align}
where $i^{-}_n$ denotes the negative token (incorrectly repeated) and $S^{-}_{m}$ denotes the negative token set which includes $m$ tokens. We select only the first $m$ tokens preceding the current token as negative samples, enabling the model to focus on highly correlated contextual ranges while preventing excessive noise introduction. The negative token set $S^{-}_{m}$ is formed as:
\begin{equation}
    S^{-}_{m} = \{i_{n-m-1}, i_{n-m}, ... , i_{n-1}\}.
\end{equation}
By using TCL conditional probability in \eqref{ct_condition}, the token level contrastive loss at each position $n$ is defined as:
\begin{equation}
        \mathcal{L}_{\text{ct}} = -\sum_{n=1}^{N} \alpha (1-p_{ct}(i_n|i_{<n}))^{\gamma} \log p_{ct}(i_n|i_{<n}).
        \label{ct_loss}
\end{equation}
Intuitively, minimizing the contrastive loss on this negative sample set containing the first m tokens reduces the likelihood of predicting incorrectly repeated tokens. Based on the above \eqref{focal_loss}, \eqref{next-k-loss}, and \eqref{ct_loss}, the loss of NTP-MRISeg can be defined as:
\begin{equation}
        \mathcal{L}_{\text{NTP-MRISeg}} = \mathcal{L}_{\text{next-token}} + \lambda_{1}\mathcal{L}_{\text{ct}} + \lambda_{2}\mathcal{L}_{\text{next-k-token}},
\end{equation}
where $\lambda_{1}$ and $\lambda_{2}$ are the weights of $\mathcal{L}_{\text{ct}}$ and $\mathcal{L}_{\text{next-k-token}}$ respectively.

\begin{table*}[t]
\caption{Comparisons with SOTA Method on QaTa-COV19 and MosMedData$+$. $\dagger$ Reptrsents That the Results Are Reported by The Original Paper. $\ast$ Reptrsents That The Results are Implemented by Offical Open-Source Code.}
\centering
\renewcommand{\arraystretch}{1.3}
\setlength{\tabcolsep}{6pt}
\begin{tabular}{llcccccc}
\hline
    \multirow{2}{*}{\textbf{Method}} & \multirow{2}{*}{\textbf{Backbone}} & \multirow{2}{*}{\textbf{Pub. Year}} & \multirow{2}{*}{\textbf{Text}} & \multicolumn{2}{c}{\textbf{QaTa-COV19}} & \multicolumn{2}{c}{\textbf{MosMedData+}} \\
\cline{5-6} \cline{7-8}
& & & & \textbf{Dice(\%)}$\uparrow$ & \textbf{mIoU(\%)}$\uparrow$ & \textbf{Dice(\%)}$\uparrow$ & \textbf{mIoU(\%)}$\uparrow$ \\
\hline
TransUNet$^{\ast}$\cite{kazemzadeh2014referitgame} & Hybrid & EMNLP 2014 & $\times$ & 78.63 & 69.13 & 71.24 & 58.44 \\
U-Net++$^{\ast}$\cite{zhou2019unet++} & CNN & TMI 2019 & $\times$ & 79.62 & 70.25 & 71.75 & 58.39 \\
nnU-Net$^{\ast}$\cite{isensee2021nnu} & CNN & Nat. Methods 2020 & $\times$ & 80.42 & 70.81 & 72.59 & 60.36 \\
Swin-Unet$^{\ast}$\cite{cao2022swin} & Transformer & ECCV 2022 & $\times$ & 78.07 & 68.34 & 63.29 & 50.19 \\
\hline
ConViRT$^{\ast}$\cite{zhang2022contrastive} & CNN & PMLR 2022 & $\checkmark$ & 79.72 & 70.58 & 72.06 & 59.73 \\
TGANet$^{\ast}$\cite{tomar2022tganet} & CNN & MICCAI 2022 & $\checkmark$ & 79.87 & 70.75 & 71.81 & 59.28 \\
GLoRIA$^{\ast}$\cite{huang2021gloria} & Hybrid & ICCV 2021 & $\checkmark$ & 79.94 & 70.68 & 72.42 & 60.18 \\
LViT$^{\dagger}$\cite{li2023lvit} & Hybrid & TMI 2023 & $\checkmark$ & 83.66 & 75.11 & 74.57 & 61.33 \\
RefSegformer$^{\ast}$\cite{wu2024towards} & Transformer & TIP 2024 & $\checkmark$ & 84.09 & 75.48 & 74.98 & 61.70 \\
DMMI$^{\ast}$\cite{hu2023beyond} & Transformer & ICCV 2023 & $\checkmark$ & 84.13 & 75.66 & 75.01 & 61.83 \\
LGA$^{\dagger}$\cite{hu2024lga} & Segment Anything & MICCAI 2024 & $\checkmark$ & 84.65 & 76.23 & 75.63 & 62.52 \\
RecLMIS$^{\dagger}$\cite{huang2024cross} & CNN & TMI 2024 & $\checkmark$ & 85.22 & 77.00 & 77.48 & 65.07 \\
CausalCLIPSeg$^{\dagger}$\cite{chen2024causalclipseg} & Hybrid & MICCAI 2024 & $\checkmark$ & 85.21 & 76.90 & - & - \\
SGSeg$^{\dagger}$\cite{ye2024enabling} & Hybrid & MICCAI 2024 & $\checkmark$ & 87.40 & 77.80 & - & - \\
GuideDecoder$^{\dagger \ast}$\cite{zhong2023ariadne} & Hybrid & MICCAI 2023 & $\checkmark$ & 89.78 & 81.45 & 77.75 & 63.60 \\
TGCAM$^{\dagger}$\cite{guo2024common} & Hybrid & MICCAI 2024 & $\checkmark$ & 90.60 & 82.81 & 77.82 & 63.69 \\
\hline
\textbf{NTP-MRISeg } & \textbf{Transformer} & \textbf{-} & $\checkmark$ & \textbf{91.10} & \textbf{83.66} & \textbf{79.18} & \textbf{65.54} \\
\hline
\label{tab1_comparisons_with_SOTA}
\end{tabular}
\end{table*}

\subsection{Memory-based HET Optimization}
\label{sec:HET}
The above TCL mitigates incorrect repeated token predictions and alleviates the long-tail distribution problem. However, certain difficult tokens remain challenging to predict and prone to errors. These incorrect tokens persist in the model's predictions and are difficult to correct, triggering a ``snowball" effect during inference that causes errors to accumulate continuously. To overcome this dilemma problem, we introduce the HET strategy which identifies frequently mispredicted tokens during training, stores them in memory, and uses them as hard negatives to guide the model toward correcting persistent errors in future updates. Specifically, we maintain a memory-based HET set $\mathcal{H}_{s,n}$ for each training sample $s$ as Fig. \ref{Fig2_model_overall}\textcolor{red}{d} at position $n$ in $t$ epoch:
\begin{equation}
    \mathcal{H}_{s,n}^{(t)} = \mathcal{H}_{s,n}^{(t-1)} \cup \{\hat{i}_{s,n}^{(t-1)} \mid \hat{i}_{s,n}^{(t-1)} \neq i_{s,n}\},
\end{equation}
where $\hat{i}_{s,n}^{(t-1)}$ represents the predicted token for sample $s$ at position $n$ in the $(t-1)$-th epoch, $i_{s,n}$ denotes the ground truth token at that position, and $\mathcal{H}_{s,n}^{(0)} = \emptyset$ (initialized as an empty set). For positions with prediction errors in the current epoch, we sort historical error tokens according to their error degree. The error degree $r_{s,n}$ is defined as:
\begin{equation}
    r_{s,n,j} = \text{logit}(\hat{i}_{s,n}) - \text{logit}(i_{s,n}),
\end{equation}
where $\text{logit}(\cdot)$ denotes the logit value of the token for sample $s$ at position $n$, and $j$ denotes each specific error token in the negative sample set. We then select the $l$/2 tokens with the highest error degrees as strong negative samples and the $l$/2 tokens with the lowest error degrees as weak negative samples, ensuring sufficient learning of difficult samples while maintaining stable performance on basic samples. The negative sample set $\mathcal{H}_{s,n}$ is:
\begin{equation}
    \mathcal{H}_{s,n} = \text{Top}_{l/2}(\mathcal{H}_{s,n}) \cup \text{Bottom}_{l/2}(\mathcal{H}_{s,n}).
\end{equation}
For each currently mispredicted position $(s,n)$, we construct the $\mathcal{L}_{\text{HET}}$ as:
\begin{equation}
    \mathcal{L}_{\text{HET}} = \frac{1}{|\mathcal{P}_{\text{error}}|} \sum_{(s,n) \in \mathcal{P}_{\text{error}}} \ell_{\text{HET}}(s, n),
\end{equation}
where $\mathcal{P}_{\text{error}}$ is the set of all positions with prediction errors in the current batch, and $\ell_{\text{HET}}$ for a single position is defined as:
\begin{equation}
    \ell_{\text{HET}}(s, n)\!\!=\!\! -\!\log \!\frac{\exp(\text{logit}(i_{s,n})\!)}{\exp(\text{logit}(i_{s,n})\!) \!+\!\! \sum_{j \in \mathcal{H}_{s,n}}\!\!\!\!\! \exp(\text{logit}(j)\!)}\!.
\end{equation}
Memory-based HET optimization helps the model consolidate basic performance while focusing on difficult tokens by distinguishing between the most challenging and simplest historical errors, thereby enhancing its ability to handle complex confusion scenarios.

\begin{figure*}[t]
\centerline{\includegraphics[width=1.7\columnwidth]{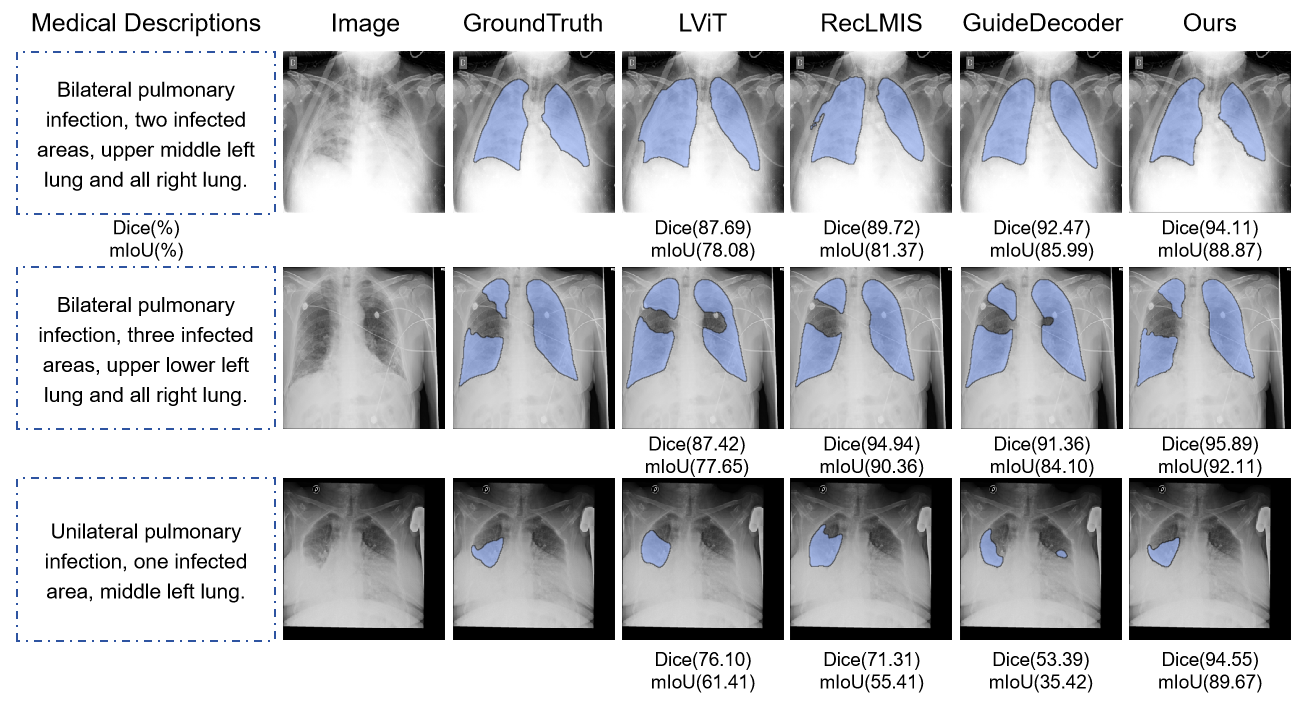}}
\caption{The visualization of the main comparison with SOTA Method on QaTa-COV19. The column titled ``Medical Descriptions" denotes the input textual referring prompt, while the column titled ``Image" signifies the input image. The column titled ``GroundTruth" represents the ground truth segmentation target. The column titled ``Ours" is the visualization result of our NTP-MRISeg. The \textcolor{blue}{blue} area is the infected segmented by NTP-MRISeg.}
\label{Fig3_CovidSOTA}
\end{figure*}

\begin{figure*}[t]
\centerline{\includegraphics[width=1.7\columnwidth]{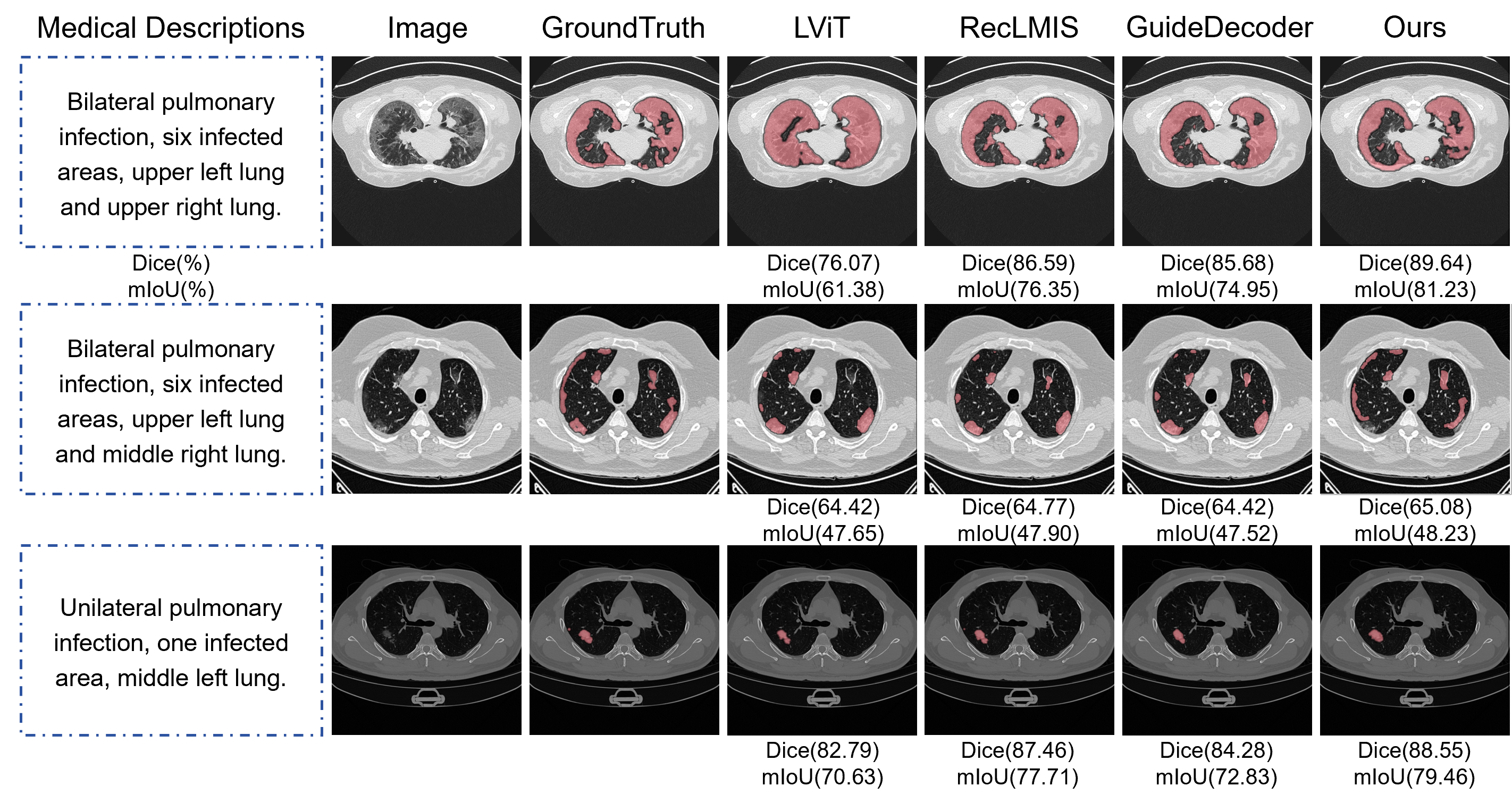}}
\caption{The visualization of the main comparison with SOTA Method on MosMedData$+$. The column titled ``Medical Descriptions" denotes the input textual referring prompt, while the column titled ``Image" signifies the input image. The column titled ``GroundTruth" represents the ground truth segmentation target. The column titled ``Ours" is the visualization result of our NTP-MRISeg. The \textcolor{red}{red} area is the infected segmented by NTP-MRISeg.}
\label{Fig3_MosSOTA}
\end{figure*}

\section{Experiments}
\subsection{Datasets and Metrics}
To comprehensively evaluate the effectiveness and robustness of our model, we conduct experiments on two MRIS datasets:
\subsubsection{QaTa-COV19 Dataset}
The QaTa-COV19 dataset\cite{degerli2022osegnet} contains 9,258 COVID-19 pneumonia X-ray radiographs. LViT\cite{li2023lvit} provides detailed medical text annotations and partitions the data for the MRIS task. The training, validation, and test sets contain 5,716, 1,429, and 2,113 images, respectively.
\subsubsection{MosMedData$+$ Dataset}
The MosMedData$+$ dataset\cite{morozov2020mosmeddata, hofmanninger2020automatic} contains 2,729 CT scan slices of lung infection. LViT\cite{li2023lvit} also provides detailed medical text annotations and partitions the data for the MRIS task. The training, validation, and test sets contain 2,183, 273, and 273 images, respectively.
\subsubsection{Evaluation Metrics}
For evaluation metrics, we use two standard metrics in medical image segmentation: Mean Intersection over Union (mIoU) and Dice Similarity Coefficient (DSC). These metrics provide complementary perspectives on segmentation accuracy and serve as standard benchmarks in medical imaging which can be described as:
\begin{equation}
    \text{mIoU} = \frac{TP}{TP + FP + FN} ,
\end{equation}
\begin{equation}
    \text{Dice} = \frac{2 TP}{2 TP + FP + FN},
\end{equation}
where $TP$, $FP$, and $FN$ represent true positives, false positives, and false negatives, respectively. Both metrics range from 0 to 1, with higher values indicating better segmentation performance. The mIoU emphasizes boundary accuracy, while Dice provides a balanced assessment that is particularly sensitive to smaller anatomical structures.
\begin{figure*}[!t]
\centerline{\includegraphics[width=1.7\columnwidth]{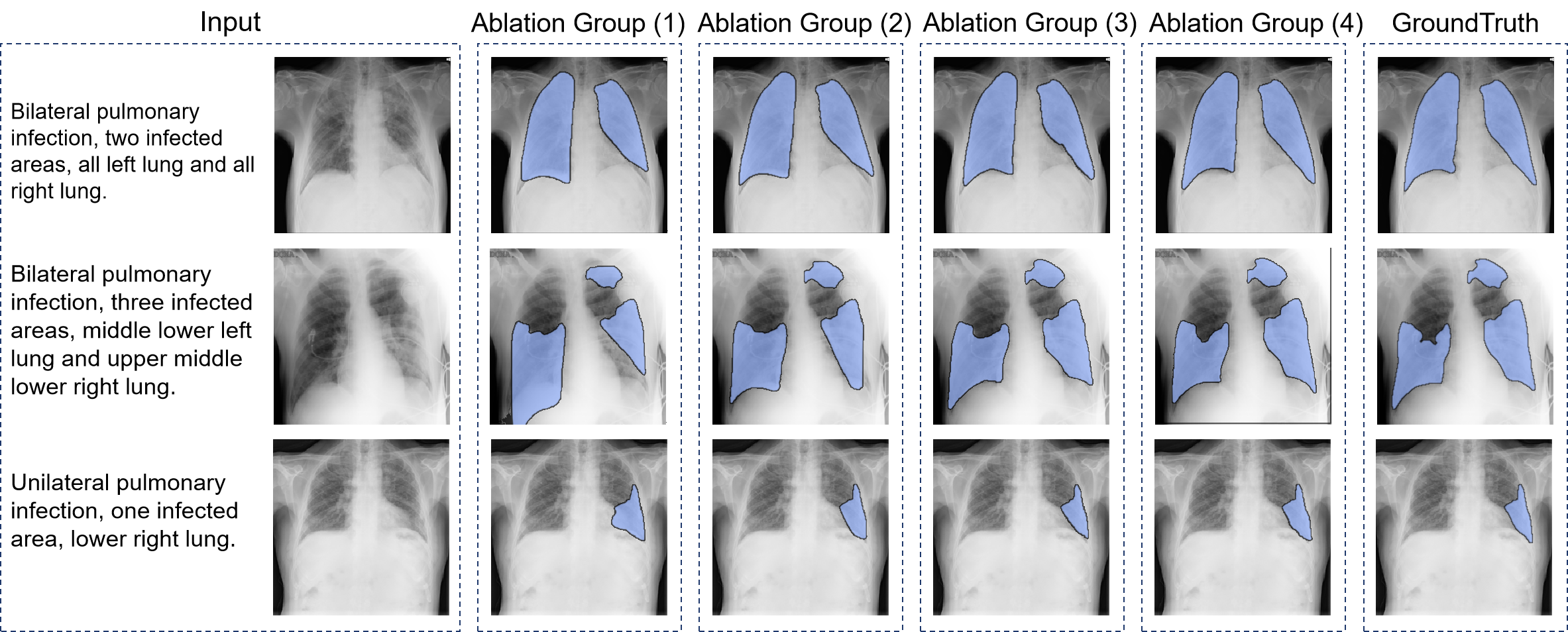}}
\caption{The visualization of the main ablation experiments. The column titled ``Input" denotes the input textual referring prompt and the input image. The column titled ``GroundTruth" represents the GT segmentation target. The column titled ``Ablation Group (1)-(4)" corresponds to the visualization results in Table \ref{cov19_ablation} of our NTP-MRISeg. The \textcolor{blue}{blue} area is the infected area segmented by NTP-MRISeg.}
\label{Fig4_ablation_study}
\end{figure*}
\subsection{Implementation Details}
We implement NTP-MRISeg under PyTorch’s distributed data parallel framework and train on 2 NVIDIA RTX A6000 Ada GPUs with 48GB memory per card. We use AdamW optimizer with a learning rate of $1 \times 10^{-4}$, weight decay 0.05, momentum parameters $\beta_{1}=0.9$ and $\beta_{2}=0.95$, dropout rate 0.1, and a WarmupCosineDecayWithMinLR with 30 steps linear warmup and cosine decay to learning rate of $1 \times 10^{-5}$ for the LoRA efficient fine-tuning. We fine-tune for 40 epochs on both QaTa-COV19 and MosMedData$+$ datasets and use beam search for mask token generation.
We conduct comprehensive ablation experiments on the challenging QaTa-COV19 and MosMedData$+$ datasets to demonstrate the effectiveness of NTP-MRISeg, which is discussed in the following 2 sections. We conduct module-by-module ablation experiments to verify the effectiveness and interactions of individual components. Additionally, we perform detailed parameter ablation studies within each module to identify optimal configurations that balance performance with computational resource consumption.
\begin{table}[!t]
\caption{Ablation Study of Proposed Components\\ on QaTa-COV19 Dataset}
\label{cov19_ablation}
\centering
\renewcommand{\arraystretch}{1.3}
\setlength{\tabcolsep}{6pt}
\begin{tabular}{c|ccccc}
\hline
 & \textbf{TCL} & \textbf{NkTP} & \textbf{HET} & \textbf{Dice(\%)}$\uparrow$ & \textbf{mIoU(\%)}$\uparrow$ \\
\hline
(1) & $\times$ & $\times$ &$\times$  & 85.23 & 74.26 \\
(2) & $\checkmark$ & $\times$ & $\times$ & 87.14 & 77.69 \\
(3) & $\checkmark$ & $\checkmark$ & $\times$ & 90.21 & 82.16 \\
(4) & $\checkmark$ & $\checkmark$ & $\checkmark$ & \textbf{91.10} & \textbf{83.66} \\
\hline
\end{tabular}
\end{table}

\begin{table}[!t]
\caption{Ablation Study of Proposed Components\\ on MosMedData$+$ Dataset}
\label{mosmed_ablation}
\centering
\renewcommand{\arraystretch}{1.3}
\setlength{\tabcolsep}{6pt}
\begin{tabular}{c|ccccc}
\hline
 & \textbf{TCL} & \textbf{NkTP} & \textbf{HET} & \textbf{Dice(\%)}$\uparrow$ & \textbf{mIoU(\%)}$\uparrow$ \\
\hline
(1) & $\times$ & $\times$ &$\times$ & 76.52 & 61.98 \\
(2) & $\checkmark$ & $\times$ & $\times$ & 77.14 & 62.79 \\
(3) & $\checkmark$ & $\checkmark$ & $\times$ & 78.23 & 64.25\\
(4) & $\checkmark$ & $\checkmark$ & $\checkmark$ & \textbf{79.18} & \textbf{65.54} \\
\hline
\end{tabular}
\end{table}

\begin{table*}[!t]
\caption{Ablation Study on NkTP $k$ Range, TCL Weight and HET Number on QaTa-COV19 Dataset}
\centering
\renewcommand{\arraystretch}{1.3}
\setlength{\tabcolsep}{13pt}
\begin{tabular}{ccccccccccccc}
\hline
    \multirow{2}{*}{\textbf{Metrics}$\uparrow$} & \multicolumn{3}{c}{\textbf{NkTP k Range ($k$)}} & & \multicolumn{3}{c}{\textbf{TCL Weight ($\lambda_{1}$)}} & & \multicolumn{3}{c}{\textbf{HET Number ($l$)}} \\
\cline{2-4} \cline{6-8} \cline{10-12}
& 8 & \textbf{16} & 32 & & 0.1 & \textbf{0.5} & 1.0 & & 30 & \textbf{50} & 80\\
\hline
\textbf{Dice(\%)$\uparrow$} & 89.64 & \textbf{91.10}  & 89.10 & & 88.75 & \textbf{91.10}  & 89.83 & & 90.55 & \textbf{91.10} & 90.27 \\
\textbf{mIoU(\%)$\uparrow$} & 81.22 & \textbf{83.66} & 80.34 & & 79.78 & \textbf{83.66} & 81.54 & & 82.73 & \textbf{83.66} & 82.27 \\
\hline
\label{Ablation_NkTP_TCL_HET}
\end{tabular}
\end{table*}

\subsection{Comparison with SOTA}
We compare our network with several mainstream CNN-based models, transformer-based models, medical segment anything based (MedSAM) segmentation models and hybrid architectures. We categorize the models based on whether they utilize text input. Table \ref{tab1_comparisons_with_SOTA} shows that medical descriptions generally improve segmentation performance, showing the necessity of the MRIS task. As shown in Fig. \ref{Fig3_CovidSOTA} and Fig. \ref{Fig3_MosSOTA}, we conducted a visualization analysis of main comparison with SOTA Method on main comparison with SOTA Method. The NTP-MRISeg we proposed achieves accurate segmentation performance. Whether on the QaTa-COV9 or MosMedData+, our model has outperformed previous SOTA.
Earlier models, such as ConViRT\cite{zhang2022contrastive} and TGANet\cite{tomar2022tganet}, employ traditional CNN structures but fail to fully utilize textual advantages due to insufficient inter-modal fusion. Recent models, including the hybrid architecture of the previous best-performing model LViT\cite{li2023lvit} and similar approaches like SGSeg\cite{ye2024enabling} and GuideDecoder\cite{zhong2023ariadne}, have achieved improved performance while following comparable architectural designs. TGCAM\cite{guo2024common} even achieved the previous SOTA with 90.60$\%$ Dice and 82.81$\%$ mIoU on QaTa-COV19 dataset and 77.82$\%$ Dice 63.69$\%$ mIoU on MosMedData$+$ dataset, respectively. Pure transformer architectures RefSegformer\cite{wu2024towards} and DMMI\cite{hu2023beyond} were also applied to MRIS, but their results were unsatisfactory due to limited adaptability to medical scenarios. Our NTP-MRISeg maintains the simplicity of pure transformer architecture while incorporating MRIS-specific optimizations such as NkTP and HET. The evaluation results on both datasets demonstrate excellent performance, achieving 91.10$\%$ Dice and 83.66$\%$ mIoU on the QaTa-COV19 dataset and 79.18$\%$ Dice and 65.54$\%$ mIoU on the MosMedData+ dataset, respectively. Particularly on the MosMedData$+$ dataset, where lesions in CT images are often more subtle than in X-ray images, our model shows sensitivity to such changes at the token level, achieving improvements of 1.36$\%$ Dice and 1.85$\%$ mIoU over the previous SOTA, respectively.

\subsection{Ablation Study}
\subsubsection{Effectiveness of Proposed Components}
As shown in Table \ref{cov19_ablation}\textcolor{red}{(1)} and Table \ref{mosmed_ablation}\textcolor{red}{(1)}, we consider the pure NTP-MRISeg transformer model without any additional modules as the baseline. mIoU was significantly improved when we introduced the TCL, as Table \ref{cov19_ablation}\textcolor{red}{(2)} shows. This improvement is attributed to effective positive and negative sample comparison that enhances the model's sensitivity, thereby eliminating model inertia. By comparing the last three rows of Table \ref{cov19_ablation} and Table \ref{mosmed_ablation}, we observe that NkTP and HET further improve mIoU performance, with more significant improvements on the QaTa-COV19 dataset (4.47$\%$ and 1.5$\%$ mIOU, respectively). This shows that NkTP effectively alleviates exposure bias, while HET plays a crucial role in helping the model handle challenging tokens.
Fig. \ref{Fig4_ablation_study} shows the visualization results of our ablation experiments on both two datasets. The results show that without TCL incorporating negative samples,  the model predicts many edge misjudgments and suffers from serious long-tail distribution problems. When NkTP and HET are introduced, incorrect predictions caused by accumulated errors are further resolved, leading to improved detail preservation and more accurate difficult area predictions.

\begin{figure}[!t]
\centerline{\includegraphics[width=\columnwidth]{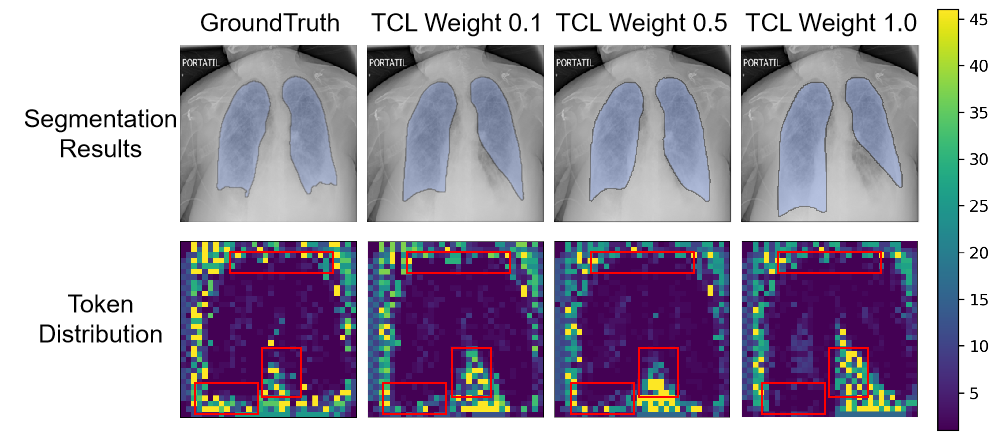}}
\caption{Heatmap visualization of token distribution according to different TCL weights (Table \ref{Ablation_NkTP_TCL_HET}). Warmer colors indicate higher frequency of prediction tokens.}
\label{Fig5_TCL_ablation}
\end{figure}

\subsubsection{Ablation Study on NkTP Range}

As described in Section \ref{sec:NkTP}, we extended NTP to NkTP to compensate for the exposure bias. However, the size of the $k$ value is very critical: if it is too small, it will not be enough to alleviate the problem of exposure bias, and if it is too large, the model will produce significant errors when predicting long-distance tokens, affecting its next-token prediction and hindering model convergence. Therefore, we selected three groups of $k$ values for ablation experiments as Table \ref{Ablation_NkTP_TCL_HET}. In general, when $k = 16$ , a good balance can be achieved, the improvement of mIoU is satisfactory, and the increase in training cost is within an acceptable range.

\subsubsection{Ablation Study on TCL Weight}
We conduct ablation experiments on the TCL weight and find that the model achieves optimal performance at a weight of 0.5 as shown in Table \ref{Ablation_NkTP_TCL_HET}. When $\lambda_1=0.1$, the weight seems insufficient to leverage the advantages of negative samples. When $\lambda_1=1.0,$ the excessive auxiliary loss disrupts the convergence of the main segmentation loss. According to the token distribution visualization in Fig. \ref{Fig5_TCL_ablation}, background regions exhibit high-frequency similar tokens while lesion areas show low-frequency unique tokens due to distinct pathological structures. When $\lambda_1=0.1$, insufficient weighting fails to leverage negative sample (background token) knowledge effectively to promote low-frequency (lesion) token prediction, resulting in excessive high-frequency token predictions and inaccurate lesion edge delineation. Conversely, when $\lambda_1=1.0$, overemphasis on negative samples causes predictions to follow distribution patterns while ignoring pathological structures, adversely affecting NTP performance.

\subsubsection{Ablation Study on HET Number}
As described in Section \ref{sec:HET}, HET is a model optimization method for difficult tokens that is introduced in the later stages of training to achieve model refinement. We need to maintain the model's learned features while improving its misperceptions based on memory. By ranking HET error levels, we can balance both less serious error tokens and the most challenging error tokens. But the participation ratio of difficult and easy samples significantly affects model refinement quality.

Since HET is memory-based, storing too many HET prevents the model from focusing on the most challenging error tokens. Conversely, adding only a small number of HETs provides insufficient samples for the model to learn useful knowledge. According to our results in Table \ref{Ablation_NkTP_TCL_HET}, when $l=50$, the model can better optimize from the memorized HET.

\section{Conclusion}
In this paper, we observe that previous medical reference segmentation models often rely on complex cross-attention structures or additional segmentation modules. Therefore, we propose NTP-MRISeg, a pure transformer-based autoregressive next token prediction model. This approach elegantly achieves language-visual feature fusion through clear input sequence construction. However, applying NTP to MRIS presents challenges including cumulative errors and task fine-grainedness. We effectively address these issues by designing a series of token-level training and optimization strategies. Our experiments on the challenging QaTa-COV19 and MosMedData$+$ datasets demonstrate NTP-MRISeg's excellent accuracy, proving that this new paradigm can be successfully adapted to MRIS tasks.

\appendices

%\section*{Acknowledgment}

{\color{black}
\bibliographystyle{IEEEtran}
\bibliography{tmi.bib}
}

\end{document}